%% file: 0-main.tex
\documentclass{article}
\usepackage{ijcai22}

\usepackage{times}
\usepackage{soul}
\usepackage{url}
\usepackage[hidelinks]{hyperref}
\usepackage[utf8]{inputenc}
\usepackage[small]{caption}
\usepackage{graphicx}
\usepackage{amsmath}
\usepackage{amsthm}
\usepackage{booktabs}
\usepackage{amsfonts}
\usepackage{mathtools}
\usepackage{amssymb}
\usepackage{booktabs}
\usepackage{mdframed}
\usepackage[table]{xcolor}
\usepackage{subcaption}
\usepackage{multirow}
\usepackage[ruled,linesnumbered]{algorithm2e}

\title{A Unified Framework for Adversarial Attack and Defense \\ in Constrained Feature Space}

\author{
Thibault Simonetto$^1$\and
Salijona Dyrmishi$^1$\and
Salah Ghamizi$^1$\and \\
Maxime Cordy$^1$\And
Yves Le Traon$^1$ \\
\affiliations
$^1$University of Luxembourg\\
\emails
\{thibault.simonetto, salijona.dyrmishi, salah.ghamizi, maxime.cordy, yves.letraon\}@uni.lu,
}

\DeclareMathOperator*{\argmax}{arg\,max}

\newcommand{\moeva}{MoEvA2}

\begin{document}

\maketitle

\begin{abstract}
The generation of feasible adversarial examples is necessary for properly assessing models that work in constrained feature space. However, it remains a challenging task to enforce constraints into attacks that were designed for computer vision. 
We propose a unified framework to generate feasible adversarial examples that satisfy given domain constraints. Our framework 
can handle both linear and non-linear constraints. 
We instantiate our framework into two algorithms: a gradient-based attack that introduces constraints in the loss function to maximize, and a multi-objective search algorithm that aims for misclassification, perturbation minimization, and constraint satisfaction. 
We show that our approach is effective in four 
different domains, with a success rate of up to 100\%, where state-of-the-art attacks fail to generate a single feasible example.
In addition to adversarial retraining, we propose to introduce engineered non-convex constraints to improve model adversarial robustness. We demonstrate that this new defense is as effective as adversarial retraining. Our framework forms the starting point for research on constrained adversarial attacks and provides relevant baselines and datasets that future research can exploit.
\end{abstract}

\input{1-introduction}
\input{3-problem_formulation}
\input{4-c_pgd}
\input{5-moeva}

\input{6-experiments_settings}
\input{7-evaluation_results}
\input{8-conclusion}

\appendix

\input{A-related_work}
\input{B-problem_formulation}
\input{C-experimental_protocol}
\input{D-evaluation_results}

\section*{Acknowledgments}

Salijona Dyrmishi is supported by the Luxembourg National Research Funds
(FNR) AFR Grant 14585105.

\bibliographystyle{named}
\bibliography{generic.bib}

\end{document}

%% file: 1-introduction.tex
\section{Introduction}
\label{sec:introduction}

Research on adversarial examples initially focused on image recognition~\cite{dalvi2004adversarial,szegedy2013intriguing} but has, since then, demonstrated that the adversarial threat concerns many domains including cybersecurity ~\cite{pierazzi2020intriguing,sheatsley2020adversarial}, natural language processing~\cite{alzantot2018generating}, software security~\cite{yefet2020adversarial}, cyber-physical systems~\cite{li2020conaml}, finance~\cite{ghamizi2020search}, manufacturing~\cite{mode2020crafting}, and more.

A peculiarity of these domains is that
the ML model is integrated in a larger software system that takes as input domain objects (e.g. financial transaction, malware, network traffic).
Therefore altering an original example in any direction may result in an example that is \emph{infeasible} in the real world. This contrasts with images that generally remain valid after slight pixel alterations. Hence, a successful adversarial example should not only fool the model and keep a minimal distance to the original example, but also satisfy the inherent \emph{domain constraints}. 

As a result, generic adversarial attacks that were designed for images -- and are unaware of constraints -- equally fail to produce feasible adversarial examples in constrained domains \cite{ghamizi2020search,tian2020exploring}. A blind application of these attacks would 
distort model robustness assessment and prevent the study of proper defense mechanisms.

Problem-space attacks are algorithms that directly manipulate \emph{problem objects} (e.g. malware code \cite{aghakhani2020malware,pierazzi2020intriguing}, audio files \cite{du2020sirenattack}, wireless signal \cite{sadeghi2019physical}) to produce adversarial examples. While these approaches guarantee by construction that they generate feasible examples, they require the specification of 
domain-specific transformations \cite{pierazzi2020intriguing}. Their application, therefore, remains confined to the particular domain they were designed for. Additionally, the manipulation and validation of problem objects are computationally more complex 
than working with feature vectors. 

An alternative to problem-space attacks is feature-space attacks that enforce the satisfaction of the domain constraints. Some approaches for constrained feature space attacks modify generic gradient-based attacks to account for constraints~\cite{sheatsley2020adversarial,tian2020exploring,erdemir2021adversarial} but are limited to a strict subset of the constraints that occurs in real-world applications (read more in Appendix A of our extended version\footnote{Extended version: \url{https://arxiv.org/abs/2112.01156}.}, where we discuss the related work thoroughly). Other approaches tailored to a specific domain manage to produce feasible examples ~\cite{chernikova2019fence,li2020conaml,ghamizi2020search} but would require drastic modifications throughout all their components 
to be transferred to other domains. To this date, there is a lack of generic attacks 
for robustness assessment of domain-specific models and a lack of cross-domain evaluation of defense mechanisms.

In this paper, we propose a unified framework\footnote{
\url{https://github.com/serval-uni-lu/moeva2-ijcai22-replication}}
for constrained feature-space attacks that applies to different domains without tuning \emph{and} ensures the production of feasible examples. Based on our review of the literature and our analysis of the covered application domains, we propose a \emph{generic constraint language} that enables the definition of (linear and non-linear) relationships between features. We, then, automatically translate these constraints into two attack algorithms that we propose. The first is \emph{Constrained Projected Gradient Descent} (C-PGD) -- a white-box alteration of PGD that incorporates differentiable constraints as a penalty in the loss function that PGD aims to maximize, and post-processes the generated examples to account for non-differentiable constraints. The second is Multi-Objective EVolutionary Adversarial Attack (\moeva{}) -- a grey-box multi-objective search approach that treats misclassification, perturbation distance, and constraints satisfaction as three objectives to optimize. The ultimate advantage of our framework is that it requires the end-user only to specify what domain constraints exist over the features. The user can then 
apply any of our two algorithms to generate feasible examples for the target 
domain.

We have conducted a large empirical study to evaluate the utility of our framework. Our study involves four datasets from finance and cybersecurity, and two types of classification models (neural networks and random forests). Our results demonstrate that our framework successfully crafts feasible adversarial examples. Specifically, \moeva{} does so with a success rate of up to 100\%, whereas C-PGD succeeded on the finance dataset only (with a success rate of 9.85\%).

In turn, we investigate strategies to improve model robustness against feasible adversarial examples. We show that adversarial retraining on feasible examples can reduce the success rate of C-PGD down to 2.70\% and the success rate of the all-powerful \moeva{} down to 85.20\% and 0.80\% on the finance and cybersecurity datasets, respectively.

%% file: 3-problem_formulation.tex
\section{Problem Formulation}
\label{sec:problem-formulation}

We formulate below the problem for binary classification. We generalize to multi-class classification problems in Appendix B.
\subsection{Constraint Language}

Let us consider a classification problem defined over an input space $Z$ 
and a binary set $\mathcal{Y} = \{0, 1\}$. Each input $z \in Z$ is an object of the considered application domain (e.g. malware \cite{aghakhani2020malware}, network data \cite{chernikova2019fence}, financial transactions \cite{ghamizi2020search}). We assume the existence of a feature mapping function $ \varphi$ : $Z \longrightarrow \mathcal{X} \subseteq \mathbb{R}^n$  that maps $Z$ to an $n$-dimensional feature space $\mathcal{X}$  over the feature set  $F = \{ f_1, f_2, ..f_n\} $. For simplicity, we assume $\mathcal{X}$ to be normalized such that $\mathcal{X} \subseteq [0, 1]^n$. That is, for all $z \in Z$, $\varphi(z)$ is an $m$-sized feature vector $x = (x_1 \dots x_n)$ where $x_i \in [0,1]$ and is the $j$-th feature. Each object $z$ 
respects some natural conditions in order to be valid. In the feature space, these conditions translate into a set of constraints over the feature values, which we denote by $\Omega$. By construction, any feature vector $x$ generated from a real-world object $z$ satisfies all constraints  $\omega \in \Omega$. 

 Based on our review of the literature, we have designed a constraint language to capture and generalize the types of feature constraints that occur in the surveyed domains. Our framework allows the definition of constraint formulae according to the following grammar:
 \begin{align*}
  \omega &\coloneqq \omega_1 \land \omega_2 \mid \omega_1 \lor \omega_2 \mid \psi_1 \succeq \psi_2 \mid f \in \{\psi_1 \dots \psi_k\}\\
    \psi &\coloneqq c \mid f \mid \psi_1 \oplus \psi_2 \mid x_i
\end{align*}

 where $f \in F$, $c$ is a constant real value, $\omega, \omega_1, \omega_2$ are constraint formulae, $\succeq \in \{<, \leq, =, \neq, \geq, >\}$, $\psi, \psi_1, \dots, \psi_k$ are numeric expressions, $\oplus \in \{+, -, *, /\}$, and $x_i$ is the value of the $i$-th feature of the original input $x$. 
 
One can observe from the above that our formalism captures, in particular, feature boundaries (e.g. $f > 0)$ and numerical relationships between features (e.g. $f_1 / f_2 < f_3$) -- two forms of constraints that have been extensively used in the literature \cite{chernikova2019fence,ghamizi2020search,tian2020exploring,li2020conaml}.




\subsection{Threat Model and Attack Objective}
\label{subsec:threat-model}

Let a function $H: \: \mathcal{X} \rightarrow \mathcal{Y}$ be a binary classifier and function $h: \: \mathcal{X} \rightarrow \mathbb{R}$ be a single output predictor that predicts a continuous probability score. Given a classification threshold $t$, we can induce $H$ from $h$ with, $H(x) = \mathbf{I}_{[\![ h(x) \ge t ]\!]}$, where $\mathbf{I}_{[\![.]\!]}$ is an indicator function, that is, $\mathbf{I}$ outputs 1 if the probability score is equal or above the threshold and 0 otherwise. 

In our threat model, we assume that the attacker has knowledge of $h$ and its parameters, as well as of $F$ and $\Omega$. We also assume that the attacker can directly modify a subset of the feature vector $x = (x_1 \dots x_m)$, with $m<n$. We refer to this subset as the set of \textbf{mutable features}. The attacker 
can only feed the 
example to the system if this example satisfies $\Omega$.


Given an original example $x$, the \textbf{attack objective} is to generate an adversarial example $x + \delta$ such that $H(x + \delta) \neq H(x)$, $\delta < \epsilon$ for a maximal perturbation threshold $\epsilon$ under a given $p$-$norm$, and $x + \delta \in \mathcal{X}_\Omega$. 
While domain constraints guarantee that an example is feasible,
(e.g. total credit amount must be
equal to the monthly payment times the duration in months), 
we limit the maximum perturbation to produce imperceptible adversarial examples.
By convention, one may prefer $p=\infty$ for continuous features, $p=1$ for binary features and $p=2$ for a combination of continuous and binary features. We refer to such examples $x + \delta$ as a \emph{constrained adversarial example}. We also name \emph{constrained adversarial attack} algorithms that aim to produce the above optimal constrained adversarial example. We propose two such attacks.





%% file: 4-c_pgd.tex
\section{Constrained Projected Gradient Descent}
\label{sec:C-PGD}

Past research has shown that multi-step gradient attacks like PGD are among the strongest attacks \cite{kurakin2016adversarial}. PGD adds iteratively a perturbation $\delta$ that follows the sign of the gradient $\nabla$ with respect to the current adversary $x_t$ of the input $x$. That is, at iteration $t+1$ it produces the input
\begin{equation}
x^{t+1} = \Pi_{x+\delta}(x^t + \alpha sgn(	\nabla_x l(\theta_h, x_t, y)))
\label{eq:pgd}
\end{equation}
where $\theta_h$ the parameters of our predictor $h$, $\Pi$ is a clip function ensuring that $x+\delta$ remains bounded in a sphere around $x$ of a size $\epsilon$ using a norm $p$, 
and $\nabla_x l$ is the gradient of loss function tailored to our task, computed over the set of mutable features. For instance, we can use cross-entropy losses with a mask for classification tasks. We compute the gradient using the first-order approximation of the loss function around $x$.

However, as our experiments reveal (see Table \ref{tab:attack} and Section \ref{sec:empirical-evaluation}), a straight application of PGD does not manage to generate any example that satisfies $\Omega$. This raises the need to equip PGD with the means of handling domain constraints. 

An out-of-the-box solution that we have experimented is to pair PGD with a mathematical programming solver, i.e. Gurobi \cite{gurobi}. Once PGD managed to generate an adversarial example (not satisfying the constraint), we invoke the solver 
to find a solution to the set of constraints 
close to the example that PGD generated (and under a perturbation sphere of $\epsilon$ size).
Unfortunately, this solution does not work out either because the updated examples do not fool the classifier anymore or the solver simply cannot find an optimal solution given the perturbation size. 

\begin{table}
  \centering
  \begin{tabular}{lll}
    Constraints formulae     & Penalty function   \\
    \hline
    $\omega_1 \land \omega_2$ & $\omega_1 + \omega_2$ \\
    $\omega_1 \lor \omega_2$ & $ \min(\omega_1, \omega_2)$ \\
    $\psi \in \Psi = \{\psi_1, \dots \psi_k\}$  & $\min(\{\psi_i \in \Psi: \mid \psi - \psi_i \mid\})$ \\
    $\psi_1 \leq \psi_2$  & $max(0, \psi_1 - \psi_2)$   \\
    $\psi_1 < \psi_2$ & $max(0, \psi_1 - \psi_2 + \tau)$   \\
    $\psi_1 = \psi_2$ & $\mid \psi_1 - \psi_2 \mid$ \\
   
  \end{tabular}
  \caption{From constraint formulae to penalty functions. $\tau$ is an infinitesimal value.}
  \label{tab:grammar-minimisation}
 
\end{table}

In face of this failure, we conclude that this gradient-based attack cannot generate constrained adversarial examples if we do not revisit its fundamentals in light of the new attack objective. We, therefore, propose to develop a new method that considers the satisfaction of constraints as an integral part of the perturbation computation.

Concretely, we define a penalty function that represents how far an example $x$ is from satisfying the constraints. More precisely, we express each constraint $\omega_i$ as a penalty function $penalty(x, \omega_i)$ over $x$ such that $x$ satisfies $\omega_i$ if and only if $penalty(x, \omega_i) <= 0$. Table \ref{tab:grammar-minimisation} shows how each constraint formula (as defined in our constraint language) translated into such a function. The global distance to constraint satisfaction is, then, the sum of the non-negative individual penalty functions, that is,  $penalty(x, \Omega) = \sum_{\omega_i \in \Omega} penalty(x, \omega_i)$.

\label{subsec:multi-objective-constrained-adversarial-attack}

The principle of our new attack, C-PGD, is to integrate the constraint penalty function as a negative term in the loss that PGD aims to maximize. Hence, given an input $x$, C-PGD looks for the perturbation $\delta$ defined as
\begin{equation}
\begin{aligned}
     \argmax_{\delta: \left\lVert \delta \right\lVert_p \leq \epsilon}\{ l(h(x+\delta),y) -\sum_{\omega_i \in \Omega} penalty(x+\delta, \omega_i) \}
\end{aligned}
\label{eq:gradient-loss}
\end{equation}
The challenge in solving (\ref{eq:gradient-loss}) is the general non-convexity of $penalty(x, \Omega)$. 
To recover tractability, we propose to approximate (\ref{eq:gradient-loss}) by a convex restriction of $penalty(x, \Omega)$ to the subset of the convex penalty functions. Under this restriction, all the penalty functions used in (\ref{eq:gradient-loss}) are convex, and we can derive the first-order Taylor expansion of the loss function and use it at each iterative step to guide C-PGD. Accordingly, \mbox{C-PGD} produces examples iteratively as follows:
\begin{equation}
\begin{aligned}
x^{t+1} = \Pi_{x+\delta}(R(x^t + \alpha sgn(\nabla_{x^t} l(h(x^t), y) \\ - \sum_{\phi_i} \nabla_{x^t} penalty(x^t, \phi_i))))
\end{aligned}
\label{eq:gradient-pgd}
\end{equation}
with $x^0 = x$, and $R$ a repair function. At each iteration $t$, $R$ updates the features of the example to repair the non-convex constraints whose penalty functions are not back-propagated with the gradient $\nabla_{x^t} l$ (if any).


 

%% file: 5-moeva.tex
\section{Multi-Objective Generation of Constrained Adversarial Examples}
\label{sec:moeva}

As an alternative to C-PGD, we propose \moeva{}, a multi-objective optimization algorithm whose fitness function is driven by the attack objective described in Section \ref{sec:problem-formulation}.

\subsection{Objective Function}
\label{subsec:minimisation-objectives}

We express the generation of constrained adversarial examples  as a multi-objective optimization problem that reflects three requirements: misclassification of the example, maximal distance to the original example, and satisfaction of the domain constraints. By convention, we express these three  objectives as a minimization problem. 

The first objective of a constrained attack is to cause misclassification by the model. When provided an input $x$, the binary classifier H outputs $h(x)$, the prediction probability that $x$ lies in class $1$. If $h(x)$ is above the classification threshold $t$, the model classifies $x$ as $1$; otherwise as $0$. Without knowledge of $t$, we consider $h(x)$ to be the distance of $x$ to class 0. By minimising $h(x)$, we increase the likelihood that the $H$ misclassifies the example irrespective of $t$. Hence, the first objective that \moeva{} minimizes is $g_1(x) \equiv h(x).$

The second objective is to minimize perturbation between the original example and the adversarial example, to limit the perceptibility of the crafted perturbations. We use the conventional $L_p$ distance to measure this perturbation. The second objective is  $g_2(x) \equiv L_p(\hat{x_0} - x_0).$

The third objective is to satisfy the domain constraints. Here, we reuse the the penalty functions that we defined in Table \ref{tab:grammar-minimisation}. The third and last objective function is thus \[g_3(x) \equiv \sum_{\omega_i \in \Omega} penalty(x, \omega_i).\]

Accordingly, the constrained adversarial attack objective translates into \moeva{} into a three-objective function to minimize with three validity conditions, that is:
 \begin{align*}
    minimise\ & g_1(x) \equiv h(x)   & s.t.\
    g_1(x) &< t\\
    minimise\ &g_2(x) \equiv L_p(\hat{x_0} - x_0)  & g_2(x) &\leq \epsilon\\
    minimise\ &g_3(x) \equiv \sum_{\omega_i} penalty(x, \omega_i)  & g_3(x) &= 0
\end{align*}
and this three-objective function also forms the fitness function that \moeva{} uses to assess candidate solutions.

  

\subsection{Genetic Algorithm}

We instantiate \moeva{} as a multi-objective genetic algorithm, namely based on R-NSGA-III~\cite{vesikar2018reference}. We describe below how we specify the different components of this algorithm. It is noteworthy, however, that our general approach is not bound to R-NSGA-III. In particular, the three-objective function described above can give rise to other search-based approaches for constrained adversarial attacks. 

\textbf{Population initialization.} The algorithm first initializes a population $P$ of $L$ solutions. Here, an individual represents a particular example that \moeva{} has produced through successive alterations of a given  original example $x$. We specify that the initial population comprises $L$ copies of $x$. The reason we do so is that we noticed, through preliminary experiments, that this initialization was more effective than using random examples. This is because the original input inherently satisfies the constraints, which makes it easier to alter it into adversarial inputs that satisfy the constraints as well.


\textbf{Population evolution.} 
\moeva{} generates new individuals with two-point binary crossover. \moeva{} selects the parents with a binary tournament over the Pareto dominance, and mutates the mutable features of the children using polynomial mutation. \moeva{} uses non-dominance sorting based on our three objective functions to determine which individuals it keeps for the next generation. We provide the details of the algorithms in Appendix B.

%% file: 6-experiments_settings.tex
\section{Experimental Settings}
\label{sec:empirical-settings}

\subsection{Datasets and Constraints}
\label{subsec:datasets-and-constraints}

Because images are devoid of constraints and fall outside the scope of our framework, we evaluate C-PGD and \moeva{} on four datasets coming from inherently constrained domains. These datasets bear different sizes, 
features, and types (and number) of constraints. We evaluate both neural networks (NN) and random forest (RF) classifiers. 
More details about datasets and models in Appendix C. 

\textbf{LCLD}
is inspired by the Lending Club Loan Data \cite{Kaggle2019}. 
Therein, examples are credit requests 
that can be accepted or rejected. 
We trained a neural network and a random forest 
that both reach an AUROC score of 0.72. Through our analysis of the dataset, we have identified constraints that include 94 boundary conditions, 19 immutable features, and 10 feature relationship constraints (3 linear, 7 non-linear). For example,
the installment (I), the loan amount (L), the interest rate (R) and the term (T) are linked by the relation  $I = L * R (1 + R)^T / ((1+R) ^T-1)$.

\textbf{CTU-13}
is a feature-engineered version of CTU-13, proposed by~\cite{chernikova2019fence}.
It includes a mix of legit and botnet traffic flows from the CTU University campus. We trained a neural network and a random forest to classify legit and botnet traffic, which both achieve an AUROC score of 0.99. We identified constraints that include 324 immutable features and 360 feature relationship constraints (326 linear, 34 non-linear). For example, the maximum packet size for TCP/UDP ports should be 1500 bytes.  

\textbf{Malware} comprises features extracted from a collection of benign and malware PE files \cite{aghakhani2020malware}. We use a standard random forest with an AUC of 0.99.
We have identified constraints that include 88 immutable features and 7 feature relationship constraints (4 linear, 3 non-linear).
For example, the sum of binary features set to 1 that describe API imports should be less than the value of features api\_nb, which represents the total number of imports on the PE file.

\textbf{URL}
comes from \cite{hannousse2021towards} and contains a set of legitimate or phishing URLs. The random forest we use has an AUROC of 0.97. We have identified 14 relation constraints between the URL features, including 7 linear constraints (e.g. hostname length is at most equal to URL length) and 7 are if-then-else constraints.

\subsection{Experimental Protocol and Parameters}

In all datasets, a typical attack would be interested in fooling the model to classify a malicious class (rejected credit, botnet, malware, and phishing URL) into a target class (accepted, legit, benign, and legit URL). 
By convention, we denote by 1 the malicious class and by 0 the target class. 

We evaluate the success rate of the attacks on the trained models using, as original examples, a subset of the test data from class 1. In LCLD we take 4000 randomly selected examples from the candidates, to limit computation cost while maintaining confidence in the generalization of the results. For CTU-13, Malware, and URL, we use respectively all 389, 1308, and 1129 test examples that are classified in class 1.

Since all datasets comprise binary, integer, and continuous features, we use the $L_2$ distance to measure the perturbation between the original examples and the generated examples.

We detail and justify in Appendices C and D the attack parameters including perturbation threshold $\epsilon$, the number of generations and population size for the genetic algorithm attack, and the number of iterations for the gradient attack.

%% file: 7-evaluation_results.tex
\section{Experimental Results}
\label{sec:empirical-evaluation}

\subsection{Attack Success Rate}
\label{subsec:rq1}

\begin{table}
\centering
\small
\begin{tabular}{lll|rrr}
\toprule
& Dataset                 & Attack &  C &  M &  C\&M \\
\midrule
\multirow{8}{*}{NN}&\multirow{4}{*}{LCLD}  &    PGD &        0.00 &             22.20 &  0.00 \\
&                        & PGD + SAT &        2.43 &              0.00 &  0.00 \\
&                        & C-PGD  &       61.68 &             22.03 &  9.85 \\
&                        & MoEvA2 &      100.00 &             99.90 & 97.48  \\
\cmidrule{2-6}
&\multirow{4}{*}{CTU-13} &    PGD &      0.00 &           100.00 &   0.00 \\
&                        &  PGD + SAT & 100.00 &              0.00 &   0.00 \\
&                        &  C-PGD  &   0.00 &              17.57 &   0.00 \\
&                        &   MoEvA2 &  100.00 &           100.00 & 100.00 \\
\midrule
\multirow{8}{*}{RF}&\multirow{2}{*}{LCLD}    & Papernot  & 0.00  & 11.86 & 0.00 \\
&                            & MoEvA2     & 99.98 & 61.84 & 41.51\\
\cmidrule{2-6}
&\multirow{2}{*}{CTU-13}  & Papernot  &  79.36 & 13.02 & 0.0  \\
&                            & MoEvA2     &  100.00&  7.62 &5.41   \\
\cmidrule{2-6}
&\multirow{2}{*}{Malware} & Papernot  & 0.00 & 51.99 & 0.00  \\
&                            & MoEvA2     & 100.00  & 100.00 & 39.30 \\
\cmidrule{2-6}
&\multirow{2}{*}{URL}  & Papernot  & 84.23 & 11.25 & 8.50 \\
&                            & MoEvA2     & 100.00 & 32.06 & 31.89 \\
\bottomrule
\end{tabular}

\caption{Success rate (C\&M) of the attacks on the neural network (NN) and random forest (RF) models, in \% of the original examples. M is the success rate disregarding constraint satisfaction; C is the ratio of original examples where the attack found examples that satisfy the constraints and are within the perturbation bound.}
\label{tab:attack}
\end{table}

Table \ref{tab:attack} shows the success rate of PGD, PGD + SAT, C-PGD, and \moeva{} on the two neural networks that we have trained on LCLD and CTU-13; and the success rate of the Papernot attack and \moeva{} on the random forests that we have trained on each dataset. More precisely, we use the extension of the original Papernot attack \cite{papernot2016transferability} that \cite{ghamizi2020search} proposed to make this attack applicable to random forests.

PGD and PGD + SAT fail to generate any constrained adversarial examples. The problem of PGD is that it fails to satisfy the domain constraints. While the use of a SAT solver fixes this issue, the resulting examples are classified correctly. C-PGD can create LCLD examples that satisfy the constraints and examples that the model misclassifies, yielding an actual success rate of 9.85\%. On the CTU-13 dataset, however, the attack fails to generate any constrained adversarial examples. The reason is that CTU-13 comprises 360 constraints, which translates into as many new terms in the function of which C-PGD backpropagates through. As each function contributes with a diverse, non-co-linear, or even opposed gradients, this ultimately hinders the attack. Similar phenomena have been observed in multi-label~\cite{song2018multi} and multi-task models ~\cite{ghamizi2021adversarial}. By contrast, \moeva{}, which enables a global exploration of the search space, 
is successful for 97.48\% and 100\% of the original examples, respectively.

MoEvA2 also manages to create feasible adversarial examples on the random forest models, with a success rate ranging from 5.41\% to 41.51\%. This indicates that our attack remains effective on such ensemble models, including with other datasets. Like PGD, the Papernot attack -- unaware of constraints -- cannot produce a single feasible example on LCLD, CTU-13, and Malware, whereas it has a low success rate (8.50\%) on URL compared to MoEvA2 (31.89\%).

\begin{mdframed}[style=MyFrame]
\textbf{Conclusion}: While adversarial attacks unaware of domain constraints  
fail, incorporating constraint knowledge as an attack objective enables the successful generation of constrained adversarial examples. 
\end{mdframed}

\subsection{Adversarial Retraining}
\label{subsec:rq2}

We, next, evaluate if adversarial retraining is an effective means of reducing the effectiveness of constrained attacks.

We start from our models trained on the original training set. We generate constrained adversarial examples (using either C-PGD or \moeva) from original training examples that each model correctly classifies in class 1. To enable a fair comparison of both methods, for the LCLD (NN), we use only the original examples for which C-PGD and \moeva{} could both generate a successful adversarial example
. In all other cases, we do not apply this restriction, since \moeva{} is the only technique that is both applicable and successful.
While \moeva{} returns a set of constrained examples, we only select the individual that maximizes the confidence of the model in its (wrong) classification, similarly to C-PGD that maximizes the model loss.

Regarding the perturbation budget, we follow established standards \cite{Carlini2019OnEA} and provide the attack with an $\epsilon$ budget 4 times larger than the defense.

\begin{table}
\centering
\small
\begin{tabular}{ll|r|r}
\toprule

Defense                     & Attack   &  LCLD &     CTU-13 \\
\midrule
None                        & C-PGD             &   9.85 &  0.00 \\
None                        & MoEvA2            &   97.48 &    100.00 \\
\midrule
C-PGD Adv. retraining *       & C-PGD             &     8.78 &    NA \\
C-PGD Adv. retraining *       & MoEvA2            &   94.90 &    NA \\
\midrule
MoEvA2 Adv. retraining *       & C-PGD             &    2.70&    NA \\
MoEvA2 Adv. retraining *     & MoEvA2            &   85.20 &  0.8\\
\midrule
Constraints augment.    & C-PGD             &   0.00 &   NA \\
Constraints augment.    & MoEvA2            &  80.43 &  0.00 \\
\midrule
MoEvA2 Adv. retrain. $\dagger$     & MoEvA2            &  82.00 &   NA\\
Combined defenses  $\dagger$       & MoEvA2            &  77.43 &  NA\\
\bottomrule
\end{tabular}
\caption{Success rate of C-PGD and MoEvA2 after adversarial retraining and constraint augmentation (on neural networks). For a fair comparison, the model denoted by the same symbols (* or $\dagger$) are trained with the same number of adversarial examples, generated from the same original samples.}
\label{tab:defense-adv-retraining}
\end{table}

\begin{table}
\centering
\small
\begin{tabular}{l|rrrr}
\toprule
Defense                  & LCLD          & CTU-13        & Malware      & URL       \\
\midrule
None                     & 41.51         & 5.41          & 39.30        & 31.89     \\
Adv. retraining    & 3.90          & 4.67          & 37.69           & 22.14        \\
Cons. augment. & 19.73         & 6.63          & 28.52         & 20.99     \\
Combined          & 0.77         & 4.67          & 28.98           & 15.94        \\

\bottomrule
\end{tabular}
\caption{Success rate of MoEvA2 on the random forest models.}
\label{tab:defense-rf}
\end{table}

Table \ref{tab:defense-adv-retraining} (middle rows) shows the results for the neural networks, and Table \ref{tab:defense-rf} (second row) for the random forests. Overall, we observe that adversarial retraining remains an effective defense against constrained attacks. For instance, on LCLD (NN) adversarial retraining using MoEva2 drops the success rate of C-PGD from 9.85\% to 2.70\%, and its own success rate from 97.48\% to 85.20\%. The fact that MoEvA2 still works suggests, however, that the large search space that this search algorithm explores preserves its effectiveness. By contrast, on CTU-13 (NN), we observe that the success rate of \moeva{} drops from 100\% to 0.8\% after adversarial retraining with the same attack.

\begin{mdframed}[style=MyFrame]
\textbf{Conclusion}: Adversarial retraining remains an effective defense against constrained adversarial attacks.
\end{mdframed}

\subsection{Defending With Engineered Constraints}
\label{subsec:rq3}

We hypothesize that an alternative way to improve robustness against constrained attacks is to augment $\Omega$ with a set of engineered constraints -- in particular, non-convex constraints. To verify this, we propose a systematic method to add engineered constraints, and we evaluate the effectiveness of this novel defense mechanism.

To define new constraints, we first augment the original data with new features engineered from existing features. Let $\hat{f}$ denote the mean value of some feature of interest $f$ over the training set. Given a pair of feature $(f_1, f_2)$, we engineer a binary feature $f_e$ 
as 
\[f_e(x) \equiv (\hat{f_1} \leq x_1) \oplus (\hat{f_2} \leq x_2) \] 
where $x_i$ is the value of $f_i$ in $x$ and $\oplus$ denotes the exclusive or (XOR) binary operator.
The comparison of the value of the input $x$ for a particular features $f_i$ with the mean $\hat{f_i}$ allows us to handle non-binary features while maintaining a uniform distribution of value across the training dataset. We use the XOR operator to generate the new feature as this operator is not differentiable. We, then, introduce a new constraint that the value of $f_e$ should remain equal to its original definition. That is, we add the constraint
\[\omega_{e} \equiv (f_e(x) =  (\hat{f_1} \leq x_1) \oplus (\hat{f_2} \leq x_2)) \]
The original examples respects these new constraints by construction.
In other words, for an adversarial attack to be successful, the attack should modify $f_e$ if the modifications it applied to $f_1$ and $f_2$ would imply a change in the value of $f_e$. 

To avoid combinatorial explosion, we add constraints only on pairs of the most important mutable features. We measure importance with the approximation of Shapley value \cite{shrikumar2017learning}, an established explainability method. 
In the end, we consider a number $M$ of pairs such that $M = \argmax_{x} {x \choose 2} < \frac{N}{4}$ where $N$ is the total number of features.

As a preliminary sanity check, we verified that constraint augmentation does not penalize clean performance and confirmed that the augmented models keep similar performance.

To evaluate this constraint augmentation defense, we use the same protocol as before, except that the models are trained on the augmented set of features. That is, we assume that the attacker has knowledge of the added features and constraints.

We show the results in Table \ref{tab:defense-adv-retraining} and \ref{tab:defense-rf} (third row). Constrained augmentation nullifies the low success rate of C-PGD on LCLD -- the gradient-based attack becomes unable to satisfy the constraints. Our defense also decreases significantly the success rate of MoEva2 in all cases except the CTU-13 random forest. For instance, it drops from 97.48\% to 80.43\% for LCLD NN, and from 100\% to 0\% on CTU-13 NN.  

\begin{figure}[]

\begin{center}
\centerline{\includegraphics[width=\columnwidth]{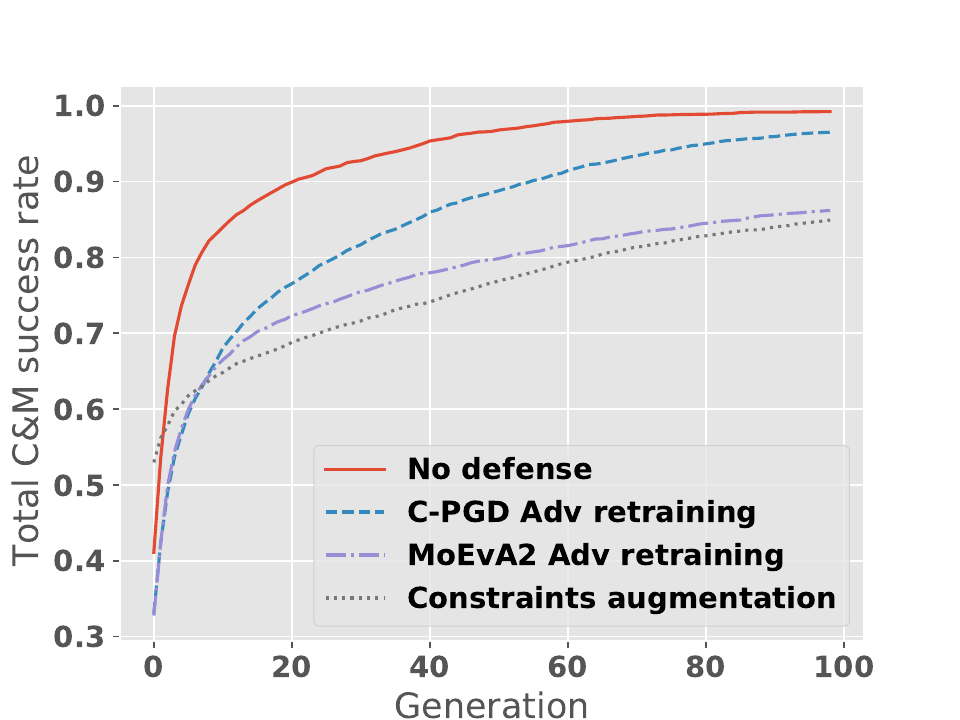}}

\caption{Success rate of \moeva{} against the original LCLD neural network and the defended counterparts, over the generations.}
\label{fig:fitness_comparison}
\end{center}

\end{figure}

We assess the effect of constraint augmentation and adversarial retraining more finely and show, in Figure \ref{fig:fitness_comparison}, the success rate of \moeva{} on the LCLD neural network over the number of generations. Compared to the undefended model, both constrained augmentation and adversarial retraining (using \moeva) lower the asymptotic success rate. Moreover, the growth of success rate is much steeper for the undefended model. For instance, \moeva{} needs ten times more generations to reach the same success rate of 84\% against the defended models than against the undefended model (100 generations versus 12 generations). Adversarial retraining using C-PGD is less effective: while it reduces the success rate in the earlier generations, its benefits 
diminishes as the \moeva{} attack runs for more generations.

\begin{mdframed}[style=MyFrame]
\textbf{Conclusion}: Constraint augmentation is an effective alternative defense against constrained adversarial attacks. The benefits of both defense mechanisms against \moeva{} are achieved as soon as the earliest generations and persist throughout the attack process. 
\end{mdframed}

\subsection{Combining Defenses}
\label{subsec:rq4}

We investigate whether the combination of constraint augmentation with adversarial retraining yields better results. A positive answer would indicate that constraint augmentation and adversarial retraining have complementary benefits.

We add to the models the same engineered constraints as we did previously. We also perform adversarial retraining on the augmented models, using all adversarial examples that MoEvA managed to generate on the training set. Then, we attack the defended models using MoEvA applied to the test set. For a fair comparison with adversarial retraining, we also apply this defense without constraint augmentation, using the same number of examples. We do not experiment with C-PGD, which was already ineffective when only one defense was used. Neither do we consider the datasets for which one defense was enough to fully protect the model.

Tables \ref{tab:defense-adv-retraining} and \ref{tab:defense-rf} (last rows) present the results. On LCLD (NN), the combined defenses drops the attack success rate from 97.48\% (on a defenseless model) to 77.23\%, which better than adversarial retraining (82.00\%) and constraint augmentation (80.43\%) applied separately. On the RFs, the combination either offers additional reductions in attack success rate compared to the best individual defense (LCLD and URL) or has negligible effects (CTU-13 and Malware).

\begin{mdframed}[style=MyFrame]
\textbf{Conclusion}: Constraint augmentation and adversarial training are two effective defense strategies that have complementary effects. Compared to their separate application, the combination can decrease the attack success rate by up to 5\%.  
\end{mdframed}

%% file: 8-conclusion.tex
\section{Conclusion}

We proposed the first generic framework for adversarial attacks under domain-specific constraints. We instantiated our framework with two methods: one gradient-based method that extends PGD with multi-loss gradient descent, and one that relies on multi-objective search. We evaluated our methods on four datasets and two types of models. We demonstrated their unique capability to generate constrained adversarial examples. In addition to adversarial retraining, we proposed and investigated a novel defense strategy that introduces engineered non-convex constraints. This strategy is as effective as adversarial retraining. We hope that our approach, algorithms, and datasets will be the starting point of further endeavor towards studying feasible adversarial examples in real-world domains that are inherently constrained.

%% file: A-related_work.tex
\section{Related work}
\label{sec:related}

\subsection{Genetic algorithm for adversarial attacks}

Genetic algorithms are considered as black-box  attack candidates by researchers in many domains. In the category of natural language processing, there exist several attacks using genetic algorithms as a search method. \cite{alzantot2018generating} were the first to use GA to replace words in a sentence with their synonyms to  generate adversarial examples. that preserve semantics and syntax. Synonyms are assumed to preserve the syntax and semantics, therefore they do not use any constraint evaluation like in \moeva{}. These attacks are specific to text and can not be used in other domains. At the same time, the scope of \moeva{} does not cover adversarial attacks in text as it considers explicit constraints between features. In parallel,~\cite{alzantot2018did} had demonstrated the first kind of adversarial attack against a speech recognition model. Similarly, they use a genetic algorithm to add noise to only the least significant bits of a random subset of audio samples. However, there was no study to show if a person could detect the fact that the audio file was tampered. The other adaptions of GA appear in~\cite{xu2016automatically}  where the authors modify malicious PDF samples to evade binary malware classifiers. Similarly, in~\cite{kucuk2020deceiving} the authors used a genetic algorithm to evade three multi-class Portable Executable (PE) malware classifiers on targeted attacks. For both PDF and PE attacks, the authors had to check the validity of the generated examples in problem space by passing them through a checker. The advantage of \moeva{} is that it ensures feasibility by including a constraint evaluation process. This avoids the need to reconstruct samples in the problem space and  evaluate them for feasibility.

\subsection{Adversarial attacks for constrained domains}
 In the constrained adversarial attacks literature, most of the studies upgrade the present attacks to support some constraints and only a few propose new algorithms tailored  to each domain specifically. One of these novel methods for crafting adversarial attacks that respect domain constraints was proposed by~\cite{kulynych2018evading}. The authors use a graphical framework where each edge represents a feasible transformation, with its weight representing the transformation cost. The nodes of the graph are the transformed examples. Despite the idea being innovative, the downside of this method is that it works only with linear models, and it comes with high computation cost. ~\cite{chernikova2019fence} builds an iterative gradient-based adversarial attack that considers groups of feature families. At each iteration, the feature of maximum gradient and its family are chosen for modification. The modifications that do not respect the constraints are updated until they enter a feasible region. The implementation provided by the authors is closely related to the botnet domain, therefore it can not be re-used in new domains.  ~\cite{li2020conaml} considers simple linear constraints for cyber-physical systems. They generate practical examples using a best-effort search algorithm. However, the solution is not scalable to practical applications that have more complex non-linear constraints. Our unified framework overcomes the barriers of domain adaptation by providing an easy language to define both linear and non-linear constraints. 

Other researchers have preferred  to upgrade existing attacks to support constraints. ~\cite{sheatsley2020adversarial} extends the traditional JSMA algorithm to support constraints resolution. They  use the concept of a primary feature in a group of interdependent features. Meanwhile, Tian et al.~\cite{tian2020exploring} introduced C-IFGSM an updated version of FGSM that considers feature correlations. They embed the correlations into a constraint matrix which is  used to calculate the Hadamard product with the sign of the gradient in order to determine the direction and magnitude of the update. ~\cite{teuffenbach2020subverting} crafted constrained adversarial examples for NIDS by extending the optimization function in Carlini and Wagner attack. They group ﬂow-based features by their modification feasibility  and assign weights to each group based on the level of difficulty of the modification.  The authors in~\cite{erdemir2021adversarial} introduce non-uniform perturbation in PGD attacks to enable adversarial retraining with more realistic examples. They use Pearson’s correlation coefficient, Shapley values, and masking to build a matrix that constrains the direction and magnitude  of change similarly to~\cite{tian2020exploring}. All the existing attacks above are upgraded to handle simple and general constraints, but do not deal with more complex domain-specific constraints. Therefore, the samples they generate are more restricted but not totally feasible. 

~\cite{ghamizi2020search} proposed a black-box, GA approach  to generate  constrained adversarial examples for an industrial credit scoring system. This work is the closest to one of the attacks we propose (\moeva).  Contrary to them, for \moeva{} we are using a multi-objective function that is not case-specific and needs neither parameter tuning nor domain expert knowledge to work. Our framework is more general and can be applied to multiple use cases without fine-tuning.

%% file: B-problem_formulation.tex
\section{Problem formulation and methods}
\subsection{Extension to multi-class classification tasks}
To simplify, we limited the definition of the problem and its solutions to binary classifiers.
We propose an extension of the problem to multi-class classification tasks and show how MoEvA2 can be extended to handle these use cases.
We consider a $n$-dimensional feature space $\mathcal{X}$  over the feature set  $F = \{ f_1, f_2, ..f_n\} $. For simplicity, we assume $\mathcal{X}$ to be normalized such that $\mathcal{X} \subseteq [0, 1]^n$.
Let $\mathcal{Y} = \{1, 2, ..., K\}$ be the set of labels of a $K$-class classification task.
Let a function $H : \mathcal{X} \rightarrow \mathcal{Y}$ be a $K$-class classifier and $h_k : \mathcal{X} \rightarrow [0, 1]$ be a single output predictor that predicts a continuous probability score of $x$ to belong to class $k$.
We can induce $H$ from $h$ with $H(x) = \argmax_{k}(\{h_k(x) | k \in \mathcal{Y}\})$.
For multi-class classification tasks, one must consider the targeted and untargeted scenarios.

Given an original example $x$ classified as $c$ and a $\mathcal{X}_\Omega$ the subset of $\mathcal{X}$ that satisfies the set $\Omega$ of domain constraints the attack objective of
\begin{enumerate}
    \item a \textit{targeted attack} towards class $\hat{y} \neq y$ is to generate an adversarial example $x+\delta$ such that $H(x+\delta) = \hat{y}$, $\delta < \epsilon$ 
    \item an \textit{untargeted attack} is to generate an adversarial example $x+\delta$ such that $H(x+\delta) \neq y$, $\delta < \epsilon$
\end{enumerate}
for a maximal perturbation threshold $\epsilon$ under a given $p-norm$, and $x + \delta \in \mathcal{X}_\Omega$.

These 2 types of attack can be solved by updating the objective functions of \moeva{}.
One can update the objective function$g_1(x)$ as follows:
\begin{enumerate}
    \item \textit{targeted attack} towards class $\hat{y} \neq y$: $g_1(x) \equiv 1- h_{\hat{y}}(x)$ (as we aim to maximize the probability $h_{\hat{y}}(x)$, we use $1- h_{\hat{y}}(x)$ to obtain a minimization towards 0 problem)
     \item \textit{untargeted attack}: $g_1(x) \equiv h_y(x)$
\end{enumerate}
while the objective functions $g_2$ (perturbation) and $g_3$ (constraints penalty) remain unchanged.

\subsection{\moeva{} Algorithm}

\begin{algorithm}
    \SetAlgoVlined
    \KwIn{$\mathbf{
    x_0}$, an original example\; 
    $G\_objectives[g_1, g_2, g_3]$, the 3 objective functions\;
    $N_{gen}$, a number of generations\;
    $L$, a population size\;
    $C$, a number of children\;
    }

    \KwOut{A population $P$ of adversarial examples minimizing the $G\_objectives$ functions\;}

    $P \leftarrow init(x_0, L)$ \;
     \For{$j = 1$ to $N_{gen}$}{
        $P_{parents} \leftarrow select\_random(P, C)$\;
        $P_{offspring} \leftarrow twoPointsCrossover(P_{parents})$\;
        $P \leftarrow P \cup polyMutate(P_{offspring})$\;
        $P_{objectives} \leftarrow evaluate(P, G\_objectives[])$\;
        $P_{survive} \leftarrow survive(P, P_{objectives}, L)$\;
    }
    \Return{$P$}
    \caption{Generation process of MoEvA2}
    \label{alg:genetic}
\end{algorithm}

Algorithm~\ref{alg:genetic} formalizes the generation process of \moeva{} described in Section 5.2.

\textbf{Population initialization.} The algorithm first initializes a population $P$ of $L$ solutions. Here, an individual represents a particular example that \moeva{} produced through successive alterations of a given  original example $x$. We specify that the initial population comprises $L$ copies of $x$. The reason we do so is that we noticed, through preliminary experiments, that this initialization was more effective than using random examples. This is because the original input inherently satisfies the constraints, which makes it easier to alter it into adversarial inputs that satisfy the constraints as well.


\textbf{Population evolution.} The algorithm proceeds iteratively and make the population evolve into new ``generations'', until it reaches a predefined number $N_{gen}$ of iterations/generations. At each iteration, \moeva{} evaluates each individual in the current population through the three-objective function defined above. Hence, we know at each stage if the algorithm has managed to successfully generate constrained adversarial examples. The evolution of the population then proceeds in three steps:

\begin{enumerate}
    \item  \textit{Crossover:} We create new individuals using two-point binary crossover~\cite{crossover}. This crossover is useful for our approach to preserve constraint satisfaction, as the ``children'' individuals keep the feature values of their ``parents''. 
    We randomly select the parents from the current population.
    
    \item \textit{Mutation:} To introduce diversity in the population, we randomly alter the features of each ``child'' (resulting from crossover) using polynomial mutation~\cite{crossover}. Our mutation operator enforces the satisfaction of constraints that involve a single feature, e.g. it preserves boundary constraints and does not change immutable features. The set of children that result from this mutation process is then added to the current population.
    
    \item \textit{Survival:}  \moeva{} next determines which individuals it should keep in the next generation. Being based on R-NSGA-III, our algorithm uses non-dominance sorting and reference directions to make this selection, based on our three objective functions. That is, we place non-dominated individuals in a first Pareto front and repeat (without replacement) until we reach $N$ Pareto fronts. Individuals in these $N$ Pareto fronts form the next generation and the others are discarded. If there are less than $L$ individuals (i.e., the population size), then, the algorithm fills the population with individuals from the $N+1$-th front, selected using reference directions -- an approach that aims to maximize
    maximizes diversity in the selection~\cite{ref_dirs_energy}.
\end{enumerate}
After the specified number of generations, the algorithm returns the last population together with the evaluation of the three objective functions.

%% file: C-experimental_protocol.tex
\section{Experimental settings}

\subsection{Dataset and models}

We evaluate C-PGD and \moeva{} on four datasets of different sizes, number of features, and types (and number) of constraints: Botnet attacks detection, credit scoring, malware detection, and URL phishing detection.

\cite{chernikova2019fence}, \cite{aghakhani2020malware} and \cite{hannousse2021towards} showed that Random Forest is the most effective model architecture to correctly classify respectively, botnet attacks, malwares, and phishing attacks on our datasets.
\cite{ghamizi2020search} demonstrated that Random Forest for a credit scoring task is among the most effective techniques. As they point out, this model architecture is particularly interesting for its performance while maintaining the interpretability, a common requirement in the financial domain.

\paragraph{Credit scoring - LCLD:}
We engineer a dataset from the publicly available Lending Club Loan Data (\textit{https://www.kaggle.com/wordsforthewise/lending-club}).
This dataset contains 151 features, and each example represents a loan that was accepted by the Lending Club.
However, among these accepted loans, some are not repaid and charged off instead.
Our goal is to predict, at the request time, whether the borrower will be repaid or charged off.
This dataset has been studied by multiple practitioners on Kaggle. 
However, the original version of the dataset contains only raw data and to the extent of our knowledge, there is no featured engineered version commonly used.
In particular, one shall be careful when reusing feature-engineered versions, as most of the versions proposed present data leakage in the training set that makes the prediction trivial.
Therefore, we propose our own feature engineering. 
The original dataset contains 151 features. 
We remove the example for which the feature ``loan status'' is different from ``Fully paid'' or  ``Charged Off'' as these represent the only final status of a loan: for other values, the outcome is still uncertain. For our binary classifier, a `Fully paid'' loan is represented as 0 and a ``Charged Off'' as 1.
We start by removing all features that are not set for more than 30\% of the examples in the training set. 
We also remove all features that are not available at loan request time, as this would introduce bias. 
We impute the features that are redundant (e.g. grade and sub-grade) or too granular (e.g. address) to be useful for classification.
Finally, we use one-hot encoding for categorical features.
We obtain 47 input features and one target feature.
We split the dataset using random sampling stratified on the target class and obtain a training set of 915K examples and a testing set of 305K.
They are both unbalanced, with only 20\% of charged-off loans (class 1). 
We trained a neural network to classify accepted and rejected loans. It has 3 fully connected hidden layers with 64, 32, and 16 neurons.
Our model achieved an AUROC score of 0.7236
Our random forest model with 125 estimators reaches an AUROC score of 0.72

For each feature of this dataset, we define boundary constraints as the extremum value observed in the training set.
We consider the 19 features that are under the control of the Lending Club as immutable. 
We identify 10 relationship constraints (3 linear, and 7 non-linear ones).

\paragraph{Botnet attacks - CTU-13:}
This is a feature-engineered version of CTU-13 proposed by~\cite{chernikova2019fence}.
It includes a mix of legit and botnet traffic flows from the CTU University campus. Chernikova et al. aggregated the raw network data related to packets, duration, and bytes for each port from a list of commonly used ports. 
The dataset is made of 143K training examples and 55K testing examples, with 0.74\% examples labeled in the botnet class (traffic that a botnet generates). Data have 756 features, including 432 mutable features. We trained a neural network to classify legit and botnet traffic. It has 3 fully connected hidden layers with 64, 64, and 32 neurons. Our model achieved an AUROC score of 0.9967. We identified two types of constraints that determine what feasible traffic data is. The first type concerns the number of connections and requires that an attacker cannot decrease it. The second type is inherent constraints in network communications (e.g. maximum packet size for TCP/UDP ports is 1500 bytes). In total, we identified 360 constraints.

In addition, we build and train a random forest classifier to detect botnet attacks. 
Our model reaches an AUROC score of 0.9925

\paragraph{Malware detection - AIMED:}

Malwares are a major threat to IT systems security. 
With the recent improvement of machine learning techniques, practitioners and researchers have developed ML-based detection systems to discriminate malicious software from benign software \cite{ucci2019survey}.
Such systems are vulnerable to adversarial attacks as shown by \cite{castro2019aimed} with the AIMED attack: they successfully evade the classifier without reducing the malicious effect of the software.
We use the dataset of benign and malicious portable executable provided in \cite{aghakhani2020malware}.
In the same paper, the authors showed that including packed and unpacked benign executables with malicious ones is less biased towards detecting the packing as a sign of maliciousness.
Therefore, we select 4396 packed benign, 4396 unpacked benign,
and 8792 malicious executables. 
As in \cite{aghakhani2020malware}, we extract a set of static features: PE headers, PE sections, DLL
imports, API imports, Rich Header, File generic. 
In total, we obtain a dataset of 17 584 samples and 24 222 features.
We use 85\% of the dataset for training and validation and the remaining 15\% for testing and adversarial generation. The trained random forest classifier reaches a test AUROC of 0.9957.

From the 24 222 features, we identify 88 immutable features based on the PE format description from Microsoft.
We also extract feature relation constraints from the original PE file examples we collected and those generated by AIMED. 
For example, the sum of binary features set to 1 that describe API imports should be less than the value of features api\_nb, which represents the total number of imports on the PE file.

\paragraph{URL Phishing - ISCX-URL2016:}

Phishing attacks are usually used to conduct cyber fraud or identity theft.
This kind of attack takes the form of a URL that reassembles a legitimate URL (e.g. user's favorite e-commerce platform) but redirects to a fraudulent website that asks the user for their personal or banking data. 
\cite{hannousse2021towards} extracted features from legitimate and fraudulent URLs as well as external service-based features to build a classifier that can differentiate fraudulent URLs from legitimate ones.
The feature extracted from the URL includes the number of special substrings such as ``www", ``\&", ``,", ``\$", "and", the length of the URL, the port, the appearance of a brand in the domain, in a subdomain or in the path, and the inclusion of ``http" or ``https".
External service-based features include the Google index, the page rank, and the presence of the domain in the DNS records.
The complete list of features is present in the replication package.
\cite{hannousse2021towards} provide a dataset of 5715 legit and 5715 malicious URLs.
We use 75\% of the dataset for training and validation and the remaining 25\% for testing and adversarial generation. The random forest model obtains an AUROC score of 0.9676.

We extract a set of 14 relation constraints between the URL features.
Among them, 7 are linear constraints (e.g. length of the hostname is less or equal to the length of the URL) and 7 are Boolean constraints of the type $if a > 0 $ then $ b > 0$ (e.g. if the number of http $>$ 0 then the number slash ``/" $>$ 0).

\subsection{Experimental setup and parameters}

We propose to study the effectiveness of MoEvA2 attack against 4 models per dataset: the first one is trained with clean samples only, the second one with adversarial examples generated on the training set using MoEvA2 in addition to the clean samples, the third by augmenting the features and constraints of the clean samples and the fourth by combining adversarial retraining and constraints augmentation.
C-PGD requires a gradient-based model which is not the case of random forests therefore we do not evaluate this attack on the random forest models.

For the LCLD and CTU-13 datasets, we reuse the same maximum perturbation threshold as for the neural network.
That is 0.05 in defense and 0.2 in attack for LCLD, and 1.0 in defense, and 4.0 in attack for CTU-13.
For Malware and URL, we use the same threshold as LCLD.
As for the original study, the budget of the attack is set to 1000 generations for CTU-13, as our preliminary study showed that the attack was not effective with a limited budget on CTU-13.
We use 100 generations for the other datasets and obtain a similar success rate as for LCLD.
We discuss the choice of these parameters in the next section.

\subsection{Implementation and hardware}
We implement \moeva{} as a framework using Python 3.8.8. 
We use Pymoo's implementation of genetic algorithms and operators.
Our models are trained using Tensorflow 2.5. Moreover, \moeva{} is compatible with any classifier that can return the prediction probabilities for a given input $x$, no matter the framework used to train it. 
For the C-PGD approach, we extend the implementation of PGD proposed by Trusted AI in the Adversarial Robustness Toolbox \cite{art2018}.
We also extend the Papernot attack implementation from the safe toolbox to support random forests.
We run our experiments with neural networks on 2 Xeon E5-2680v4 @ 2.4GHz for a total of 28 cores with 128 GB of RAM.
The ones with random forests are run on 2 AMD Epyc ROME 7H12 @ 2.6 GHz  for a total of 128 cores with 256GB of RAM.

%% file: D-evaluation_results.tex
\subsection{Hyper-parameters evaluation}
\label{app:hp-params}

Our approaches present a large number of hyper-parameters.
We hypothesize that two of them have a direct impact on the success rate.
First, as commonly admit, the maximum allowed perturbation $\epsilon$.
Given a large enough perturbation $\epsilon$, an attack shall always be able to find an adversarial, as even a legit input from the target class becomes an ``adversarial''.
Second, we hypothesize that the number of iterations given to the C-PGD attack impacts the success rate.
Concerning MoEvA2, and as commonly admitted for genetic algorithms, the number of generations shall have an impact on the success rate of the attack.
Preliminary experiments gave us an approximation of what would be a good value for these parameters. 
We chose an $\epsilon$ value of $0.2$ and $4$, respectively, for the LCLD and CTU-13 projects. Note that the maximum perturbation for normalized features between $[0,1]$ and $L_2$ distance is $\sqrt{n}$ where $n$ is the number of features, which is $6.86$ and $27.48$ for LCLD and CTU-13 respectively.
With \moeva{} attack, we use $100$ and $1000$ generation for LCLD and CTU-13 respectively. 
We started by using the same budget for both datasets.
We found on a trial run on CTU-13 that, with only $100$ generation, the objective function $g_1$ (misclassification probability) was slowly decreasing over the generation, but was not able to reach the threshold.
Therefore, we attempted to augment the budget to $1000$.
We propose to study the impact of $\epsilon$ and the number of generations/iterations on the success rate $C\&M$.
We study the variation of the success rate of our four attacks for $\epsilon \in [0.025, 0.5, 0.1, 0.2 ,0.4]$ (LCLD) and $\epsilon \in [0.5, 1, 2, 4 ,8]$ (CTU-13) with the aforementioned number of generations/iterations.

\begin{figure}[ht!]

\begin{center}
\centerline{\includegraphics[width=\columnwidth]{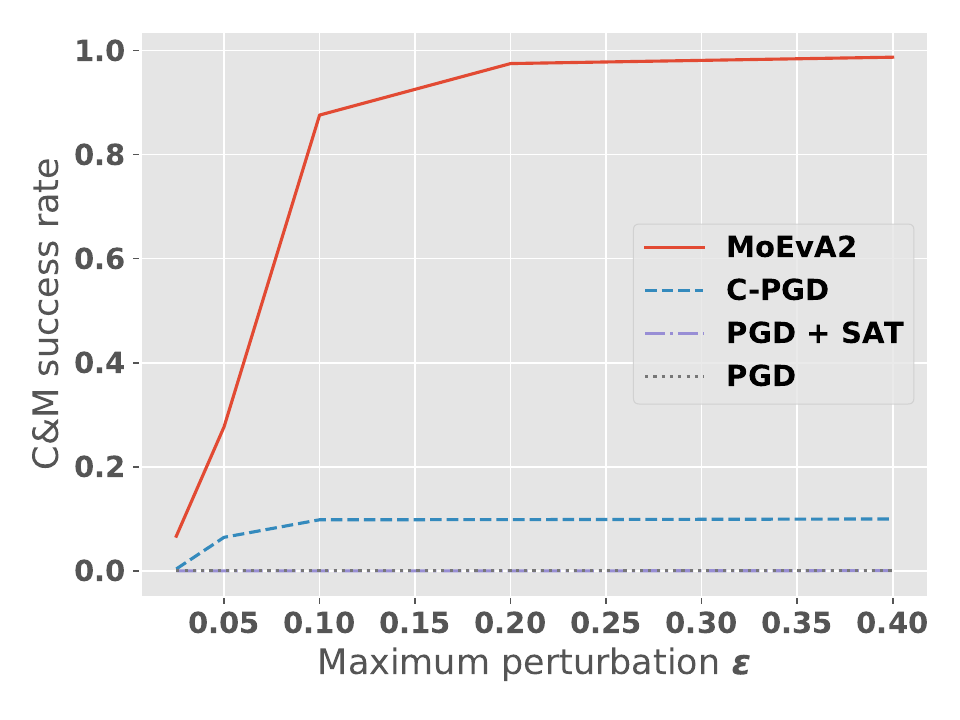}}

\caption{Success rate M\&C of different attacks over $\epsilon$ budget on LCLD. The curves for PGD and PGD+SAT overlap on the value 0.}
\label{fig:lcld_eps}
\end{center}
\end{figure}

Figure \ref{fig:lcld_eps} shows the success rate of our four attacks for different $\epsilon$ budgets on the LCLD use case.
We see that C-PGD reaches a plateau for $\epsilon = 0.1$, while \moeva{} does not show any significant improvement after $\epsilon = 0.2$.
We want to have a high success rate on both attacks for our main experiments to properly assess the impact of our defense methods.
Therefore, we chose $\epsilon = 0.2$.
For the CTU-13 use case, we obtained similar results for all $\epsilon$ and for both C-PGD and \moeva{}.

We also study the impact of the number of generations/iterations on the success rate, with  $\epsilon$ fixed to the aforementioned values.

\begin{table}
\centering
\begin{tabular}{lr|r}
\toprule
& Number of Generations &  C\&M \\
\midrule
\multirow{3}{*}{LCLD }  &   50 &        95.45  \\
                        & 100 &        97.48 \\
                        & 200  &       98.45 \\      
\midrule
\multirow{3}{*}{CTU-13} &    100 &       12.92 \\
                        &  500  &   99.48 \\
                        & 1000 &   100.00 \\
\bottomrule
\end{tabular}

\caption{Success rate (C\&M) in \% of \moeva{}, for different numbers of generations.}
\label{tab:nb_gen}
\end{table}

Table \ref{tab:nb_gen} shows that \moeva{} requires different budgets depending on the use case to reach a similar success rate.
Additionally, we observe a high success rate from $50$ generations and from $500$ generations for LCLD and CTU-13 respectively.
To properly assess the effectiveness of the defense, and not be limited by the number of generations, we chose to keep $100$ and $1000$ values as our default budget throughout the paper.

\begin{table}[t]
\centering
\begin{tabular}{lr|r}
\toprule
& Number of Generations &  C\&M \\
\midrule
\multirow{3}{*}{LCLD }  &   500 &        8.88  \\
                        & 1000 &        9.85 \\
                        & 2000  &       10.75 \\      
\midrule
\multirow{3}{*}{CTU-13} &    500 &       0.00\\
                        &  1000  &   0.00 \\
                        & 2000 &   0.00 \\
\bottomrule
\end{tabular}

\caption{Success rate (C\&M) in \% of C-PGD, for different numbers of iterations.}
\label{tab:nb_iter}
\end{table}

Table \ref{tab:nb_iter} shows that the effect of the number of iterations given to C-PGD does not influence significantly the success rate of the attack.
Moreover, we see that no matter the iteration budget, C-PGD is incapable of generating a single adversarial example against the CTU-13 model.
The same experiment (omitted from the table) reveals that the classic PGD and PGD coupled with a SAT solver fail to generate a single constrained adversarial example against both models no matter the number of iterations.





\subsection{Combining constraints engineering and adversarial retraining to defend against search-based attacks.}
\label{subsec:rq4-supp}

Our previous results imply that both adversarial retraining using MoEva2 and constraint augmentation improve the robustness of CML models. We argue that the two mechanisms are complementary and can be combined for improved robustness.

To evaluate the effectiveness of combining constraints engineering with adversarial retraining to defend our model, we compare the robustness of 4 defense scenarios against \moeva{} attack: the first one is no defense and is equivalent to Section 7.1. The second approach uses adversarial retraining, the third approach uses constraint augmentation, and the last is adversarial retraining on a constraint-augmented model. 

For a fair comparison, both the constraint-augmented and the original model are adversarially retrained with the same amount of inputs, even if the success rate of MoEva2 on the constrain-augmented model is significantly lower, and thus generates fewer adversarial examples to train with.
Using this protocol, we use 94K examples generated by MoEvA2 to retrain.
That is about 3 times more than in Section 7.2.
Therefore, we expect the success rate of the attack to be different from  the success rate in Section 7.2 even though we use the same algorithm to generate the adversarial examples.

Table \ref{tab:defense-constraints-adv} presents the results for the credit-scoring task. We show in Section 7.2, that constraint augmentation alone is sufficient to protect the model against botnet detection adversarials, a combination of the two defenses is therefore superfluous.

\begin{table}[t!]
\centering
\begin{tabular}{llrrr}
\toprule
        Defense  &       C &      M &    C\&M \\
\midrule
           None                   & 100.00 & 99.90 & 97.48 \\
Augment          & 100.00 & 93.33 & 80.43 \\
         MoEvA2                   & 100.00 &  89.00 &  82.00 \\
MoEvA2 +  Augment & 100.00 &  89.23 & 77.23 \\
\bottomrule
\end{tabular}
\caption{Success rate (\%) of \moeva{} against different defense strategies according to 3 objectives, C for constraints satisfaction, M for misclassification, and C\& M for both constraints satisfaction and misclassification for the same generated example.}
\label{tab:defense-constraints-adv}
\end{table}

Starting from a constrained success rate of 97.48\% on a defenseless model, the adversarial retraining lowers it to 82\%
 while attacks against a constraint augmented model yield an 80.43\% success rate. Combining adversarial retraining on top of a constraint augmentation defense leads to a success rate of 77.23\%, improved from using only one or the other technique. 
 
\begin{mdframed}[style=MyFrame]
Conclusion: Combining constraint augmentation and adversarial retraining reduces the success rate of constrained adversarial attacks MoEva2 by about 20\% compared to unprotected models.  
\end{mdframed}

%
%


\subsection{Evaluation of Random Forest Classifiers}

Table \ref{tab:defense-adv-retraining-rf} shows the success rate of MoEvA2 against 3 models (clean, adversarially retrained and constraints augmented).
First, we observe that \moeva{} successfully generates adversarial examples for all 4 datasets on the clean random forest, although we noticed that the success rate is significantly lower for the CTU-13 dataset.

\begin{table}[t!]
\small
\centering
\begin{tabular}{l|rrrr}
\toprule

Defense                  & LCLD          & CTU-13        & Malware      & URL       \\
\midrule
None                     & 41.51         & 5.41          & 39.30        & 31.89     \\
Augment & 19.73         & 6.63          & 28.5         & 20.99     \\
\midrule
MoEvA2   & 3.90          & 4.67          & 37.69           & 22.14        \\
MoEvA2 +  Augment        & 00.77         & 4.67          & 28.98           & 15.94        \\

\bottomrule
\end{tabular}
\caption{Success rate MoEvA2 after adversarial retraining and constraint augmentation.}
\label{tab:defense-adv-retraining-rf}
\end{table}